# DocTalk: Enhancing Medical Knowledge Retrieval via Fine-tuned Large Language Models with Chain-of-Thought Reasoning for Robust Clinical Decision Support

Asad Gohar, Md Ariful Islam Mozumder, Moon-Il Joo, Hee-Cheol Kim
Institute of Digital Anti-Aging HealthCare, Inje University, Gimhae 50834, Republic of Korea;
Corresponding Author: heeki@inje.ac.kr

## Abstract:-

**Objective:** The rapid advancement of large language models (LLMs) in healthcare has opened new possibilities for AI-assisted clinical decision-making; however, safe and effective deployment remains challenging becasue of hallucination and domain misalignment. This study introduces DocTalk, a doctor-assistive question-answering (QA) system designed to generate concise, chart-grounded responses from de-identified discharge summaries.

**Methods:** We developed a high-fidelity data generation pipeline that produces 500 clinically relevant QA pairs through sentence-level decomposition, chain-of-thought-guided question formulation, and expert review. Six open-source LLMs, including three general-purpose and three biomedical models, were evaluated before and after parameter-efficient instruction tuning using Low-Rank Adaptation (LoRA). The framework integrates prompt engineering, in-context learning, and medically aligned evaluation to support factual and interpretable responses.

**Results:** On DocTalk-QA, instruction tuning produced consistent improvements across lexical, semantic, and clinical fidelity metrics. The best-performing tuned model, Llama-3-Med42-8B, achieved ROUGE-L=0.40, BERTScore-F1=0.76, and UMLS-F1=0.49. To examine external generalization beyond the DocTalk-QA dataset, the two best-performing models, Llama-3.1-8B-Instruct and Llama-3-Med42-8B, were evaluated on PubMedQA before and after instruction tuning. Accuracy improved from 0.931 to 0.977 for Llama-3.1-8B-Instruct and from 0.935 to 0.943 for Llama-3-Med42-8B.

**Conclusion:** DocTalk was deployed as a secure web-based platform for interactive EHR querying. These findings suggest that clinically grounded instruction tuning can improve medical QA performance and support practical clinical decision support.

## 1. Introduction

In recent era, a significant shift towards digital systems for managing medical systems has been observed. The electronic health records (EHRs) are most important, which have information of patient's medical history, including diagnosis, medications, procedures, laboratory test results, and follow-up plans [1]. However, there are vast amount of unstructured medical data, posing challenges for efficient information extraction and using it for the right purpose. These types of documents are generated using complex language, which is difficult to understand for the patients. Therefore, this problem can be solved by generative artificial intelligence (AI) techniques, particularly large language models (LLMs), by automatic retrieval of structured clinical information from unstructured medical reports [2-5]. Recent LLMs have made substantial advancements in medicine, demonstrating admirable performance in answering medical questions [6-8].

The integration of Artificial Intelligence (AI) in healthcare has transformed personal healthcare, enabling advancements in diagnostic accuracy [9], treatment personalization [10], [11], and patient management [12-15]. At the core of this transformation are LLMs, which show exceptional abilities in understanding and generating human-like text. Recently, in healthcare sector, it has become more capable at a growing range of language tasks, which inherently constitute a powerful tool that expands equitable to medical knowledge [16].

LLMs are getting much better across a diverse spectrum of linguistic tasks, and they inherently represent medical information more accessible to everyone. Because of this, people are getting interested in using AI tools based on LLMs in different medical tasks. Such applications encompass diagnosis assistance, the automation of administrative functions, and even helping healthcare professionals take notes. Furthermore, before we can use LLMs in real medical situations, these models need to be checked to make sure they are correct, fair, and safe for patients [17-19]. Nevertheless, prior to the development of LLMs for real medical situations, models designated for healthcare integration must unequivocally demonstrate accuracy, impartiality, and patient safety. A foundational step in the successful evolution of healthcare-specific LLMs necessitates their rigorous evaluation against established medical benchmarks. This evaluation process requires the assessment of performance on multiple-choice questions from an extensive array of sources to determine their accuracy on various tasks and medical topics [20-22]. Much of the existing research has focused on enhancing LLM-based question-answering (QA) capabilities, which are pivotal for delivering timely and accurate responses to patient inquiries. Datasets and benchmarks such as MedQA [23, 24], emrQA [25], MedMCQA [26], PubMedQA [27], and the MMLU [28] have been developed to evaluate the QA abilities of LLMs in healthcare.

There are several research studies primarily focused on clinical QA. Systems designed to summarize information for patients, like discharge instructions or brief hospital summaries, are easy for the public to understand. However, they are not effective at answering the detailed questions clinicians require, potentially causing important information to be missed [27]. Moreover, QA systems tested with exam-style questions often use multiple-choice answers or assess explanation quality. These tests typically do not mimic real EHR environments and are often limited to a few medical fields [29]. LLMs for clinical use, trained with specific instructions, perform well on general medical tasks. However, they may encounter difficulties with out-of-distribution (OOD) clinical notes that vary across institutions or templating formats, and must meet strict safety and bias standards [30]. Nevertheless, in recent studies, LLMs evaluated on medical benchmarks showed significant improvement in performance after instruction-tuning or fine-tuning based on few-shot and chain-of-thought (CoT) strategic prompting techniques [31-33]. The models are trained using these techniques to better understand instruction like humans, and produce truthful reasoning in a step-by-step thinking process [34-36]. Therefore, performance optimization of pretrained LLMs can achieve state-of-the-art (SOTA) results on medical QA benchmarks with some specific fine-tuning techniques.

In this study, we present the DocTalk prompting platform, a doctor-assistive QA system that is directly based on actual and de-identified discharge and EHR documentation. We processed the clinical-oriented data by splitting the notes into atomic sentences, writing doctor-style questions with a brief CoT prompt. To carry out this analysis, we used six LLMs (i.e., 3 general and 3 biomedical) for evaluation, and each of them is adapted with low-rank adaptation (LoRA) fine-tuning technique [37]. DocTalk aims to relieve the burden on physicians, improve accuracy, and able to scale to non-homogeneous real-world environments by decreasing the amount of time clinicians spend searching charts, standardizing the quality of answers, and basing the results on existing clinical ontologies. The proposed clinical-assistive QA system can return concise and chart-grounded answers from real discharge/EHR notes, optimized for clinical decision support.

## 2. Related Works

The application of LLMs in healthcare has rapidly emerged as a central component in modern healthcare AI systems, particularly for clinical QA, information retrieval, and decision support [38-40]. Prior research in this area can be broadly categorized into five streams: 1) general LLM applications in healthcare, 2) clinical QA over EHR narratives, 3) instruction-tuned medical LLMs, 4) prompting and parameter-efficient adaptation strategies, and 5) biomedical QA benchmarks and foundation models.

### 2.1. LLMs in Healthcare

Recently, it has been analyzed that LLMs can generate coherent, context-aware, and clinical-style text, making them attractive for healthcare applications. Early studies like GPT-4 and its successors showed strong performance across diverse medical reasoning tasks, including diagnosis-related questions, treatment planning, and clinical documentation, exceeding human-level performance on standardized benchmarks [41]. However, subsequent analysis revealed important limitations, including skewed memorization effects, factual inconsistency, and vulnerability to distributional shifts in clinical data [42]. To address these challenges, researchers have explored domain-specific pretraining on EHRs. Models trained directly on longitudinal clinical data have demonstrated improved contextual understanding of patient histories and procedural timelines compared to general-purpose LLMs [43]. Despite these advances, deploying LLMs in real-world clinical settings remains difficult due to safety concerns, interpretability requirements, and the highly heterogeneous nature of EHR documentation across institutions.

### 2.2. Clinical Question Answering over EHR Narratives

Clinical QA over EHR narratives differs fundamentally from exam-style or patient-facing QA tasks. Early datasets such as emrQA introduced large-scale question-answer pairs derived from electronic medical records, enabling research on extracting answers from unstructured clinical notes [44]. Traditional sequence models, including LSTM-based architectures, were initially applied to these tasks but struggled with long-context reasoning and document-level coherence [45]. In [46], they mentioned that recent approaches have focused on retrieval-augmented and experience-based QA systems, which influence past patient records to answer discharge-related questions. Moreover, many existing QA systems prioritize patient discharge summarization rather than clinician-oriented decision support, potentially omitting nuanced medical details required by healthcare professionals.

### 2.3. Instruction-Tuned Clinical LLM

In medical tasks, instruction-tuning has emerged as a powerful strategy for adapting LLMs. Models like Med-

PaLM demonstrated that targeted instruction fine-tuning can significantly enhance medical reasoning, diagnostic accuracy, and explanation quality [47]. The following work showed that LLMs encode substantial latent clinical knowledge even before fine-tuning, which can be unlocked through carefully designed instructions and domain-specific supervision [35]. More recent medical LLMs, such as LlamaCare and other healthcare-specialized variants, have explored large-scale instruction tuning for medical knowledge sharing and clinician assistance [48]. Although these models perform well on curated benchmarks, they often rely on exam-style data and may not generalize better to real-world EHR narratives.

### 2.4. Prompting Strategies and Parameter-Efficient Adaptation

Prompting-based technique, which offers an alternative to fine-tuning by guiding model behavior through structured inputs. In LLMs, CoT prompting has been shown to produce step-by-step reasoning, improving performance on complex tasks, including medical QA [49]. Self-consistency and multi-sample reasoning further enhance reliability by aggregating multiple reasoning paths [50]. Concurrently, parameter-efficient fine-tuning techniques such as Low-Rank Adaptation (LoRA) enable efficient specialization of large models by updating only a small subset of parameters [51, 52]. Therefore, these techniques significantly reduce computational overhead while preserving model performance.

### 2.5. LLMs in Biomedical Question Answering

Biomedical QA benchmarks have played a crucial role in evaluating language understanding and reasoning in medical contexts. Foundational models, such as BERT, GPT, and T5, established transfer learning paradigms that were later extended to biomedical text through domain-adaptive pretraining [53-55]. Large open-source models, including OPT, BLOOM, PaLM, and LLaMA, further expanded the scale and accessibility of foundation models for biomedical research [56-59]. Recently, the researchers focused on generalist biomedical AI systems capable of reasoning across clinical, molecular, and literature-based tasks [60]. Models such as BioMistral have shown improved performance on biomedical language tasks through targeted pretraining on medical corpora [61].

In contrast to prior research works, our study bridges the gap between benchmark-driven biomedical QA and real-world clinical decision support. By constructing a clinician-verified QA dataset directly from de-identified EHR discharge summaries and integrating CoT reasoning with parameter-efficient instruction tuning, our proposed system, DocTalk-QA, explicitly targets the needs of healthcare professionals. Unlike exam-style QA or patient-oriented summarization systems, our approach emphasizes concise, chart-grounded answers that preserve clinical accuracy and interpretability.

## 3. Materials and Methods

The overall workflow of the proposed DocTalk framework is illustrated in Figure 1. The framework consists of five major components: construction of the DocTalk-QA dataset from de-identified discharge/EHR notes, unified prompting and hallucination-mitigation strategies, LoRA-based instruction tuning of general-purpose and biomedical LLMs, internal and external evaluation using DocTalk-QA and PubMedQA, and final deployment through a secure DocTalk web platform. This design enables the system to generate concise, chart-grounded answers while supporting both in-domain clinical QA evaluation and out-of-domain biomedical benchmark validation.

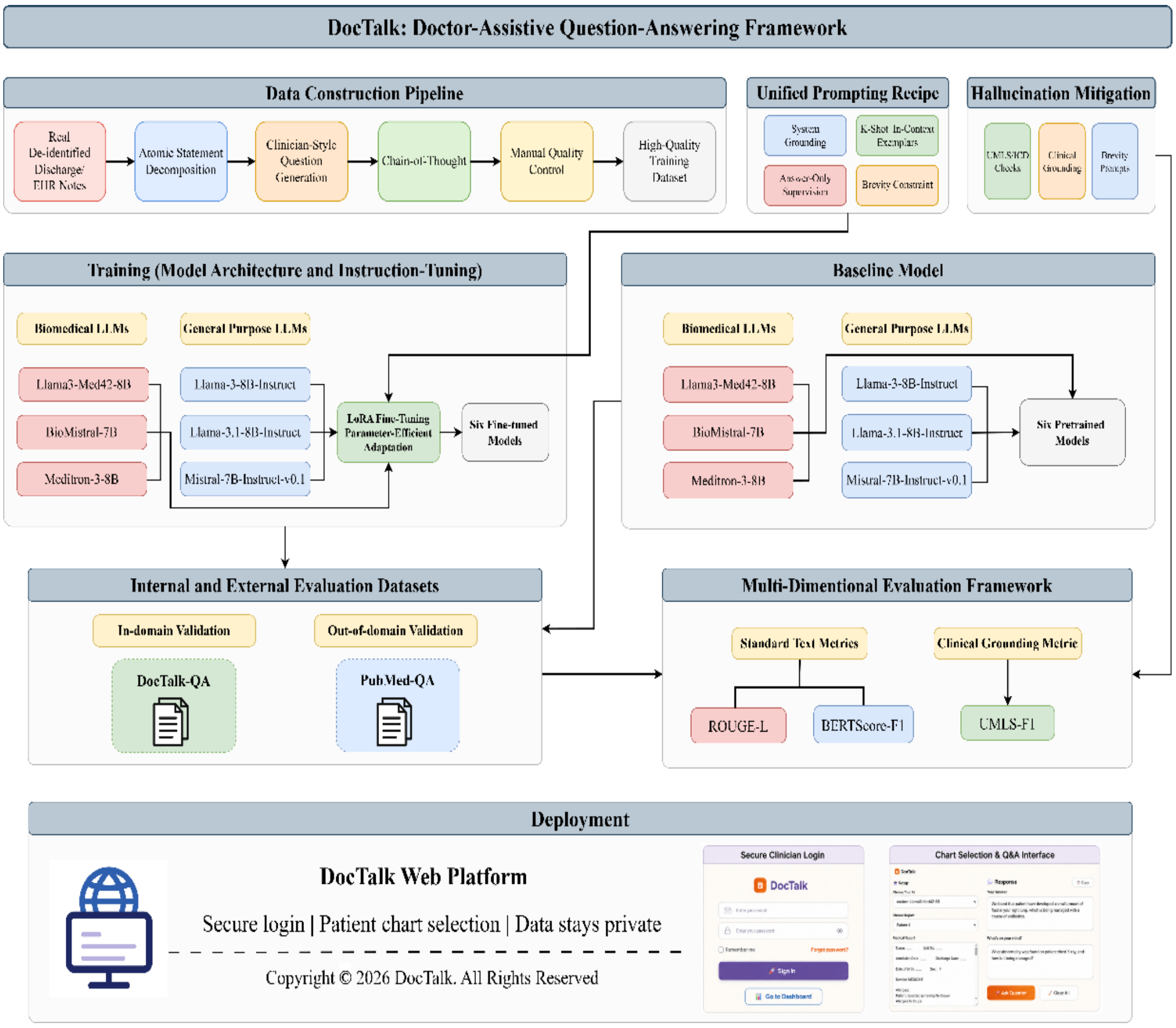

**Figure 1:** Overall workflow of the DocTalk doctor-assistive question-answering framework.

### 3.1. Datasets and Tasks

In this study, two different datasets were utilized for the development and evaluation of our DocTalk QA system. MIMIC-IV [62], which is a publicly available de-identified electronic health record (EHR) dataset compiled by the Beth Israel Deaconess Medical Center (BIDMC) in Boston, in collaboration with the Massachusetts Institute of Technology (MIT) laboratory for computational physiology. PubMedQA [63] is a biomedical QA dataset that is to answer research questions by reasoning over the corresponding abstract as context. We used PubMedQA as an external benchmark to test the generalization of our model on biomedical research questioning tasks. This allowed us to compare model performance across different document types and reasoning settings.

### 3.2. Data Pre-processing and Tokenization

All the raw discharge summaries downloaded from the MIMIC-IV database were pre-processed to ensure textual consistency, clinical accuracy, and removal of irrelevant artifacts. The workflow of data pre-processing consists of format standardization, artifact removal, and medical term normalization, which can be expressed as:

$$Clean_{text} = Norm_{medica_terms}\left(Remove_{artifacts}\left(Standardize_{format}(raw_text)\right)\right) \quad (1)$$

The raw text was standardized to correct irregular text structures, restore line breaks between sections, and unify inconsistent spacing. After standardizing the raw text, irrelevant metadata were eliminated to retain only the clinically meaningful text required for QA generation. Finally, we applied the normalization technique to maintain consistency of medical terminology across the dataset.

Furthermore, tokenization was performed on the normalized data to convert it into tokens using a biomedical-specific tokenizer. We removed rare or noisy words based on a frequency threshold to ensure that only clinical terms were retained. The resulting vocabulary $V$ was defined as:

$$V = \{token | token \in medical_{corpus}, frequency(token) \geq threshold\} \quad (2)$$

In addition, special tokens such as <unk>, <pad>, and <eos> were added to manage unknown words and sequence boundaries. This tokenization process prepared the text for model training while preserving the medical meaning of important terms and improving computational efficiency.

### 3.3. DocTalk-QA Dataset from MIMIC-IV

We constructed a high-quality clinician-style QA dataset using the pre-processed text generated from discharge summaries of the MIMIC-IV database. A representative discharge summary and CoT prompting technique is shown in Figure 2, illustrating the type of rich narrative content used for question generation. To create the dataset, we followed a multiple-step, CoT-assisted prompting pipeline (Figure 3), designed to enforce stepwise clinical reasoning during both question generation and answer formulation. Each discharge summary was first segmented into semantically coherent sentences to maintain contextual integrity. We used *Meta's Llama-3.3-70B-Instruct* model for this task. This particular model was selected based on its linguistic adherence to clinical phrasing and section conventions in discharge notes, its independence of instruction on schema-constrained outputs, and its ability to handle context (i.e., capability of 128K tokens). Compared to traditional sentence splitters such as NLTK, the LLM-based method demonstrated higher accuracy, especially in handling irregular formatting and placeholders introduced during anonymization. To ensure integrity, the LLM was instructed to preserve original sentence structures, and outputs were string-matched against the source for verification. Then, using prompt engineering, the model generated clinically relevant questions aligned with the reasoning style of healthcare professionals.

The QA pairs covered six major clinical categories, which include diagnosis (17.8%), medications (18.6%), procedures and tests (20.4%), treatment and hospital course (21.9%), symptoms and complications (16.3%), and post-discharge care and follow-up (5.0%). To ensure reliability and clinical safety, each QA pair was reviewed by a licensed medical expert. They were assessed for clinical correctness, answer completeness, safety (i.e., absence of misleading or potentially harmful information), and alignment with real-world clinician information needs. Moreover, we applied ICD-10 coding checks for diagnostic accuracy and UMLS-based validation to confirm biomedical concept grounding. This quality assurance process produced a curated set of 500 high-quality QA pairs, each pair embeds traceable CoT reasoning steps, facilitating interpretability and structured evaluation. The final dataset was formatted in JSON/JSONL, with each entry structured as a triplet (question, context, answer), ready for model training and inference tasks.

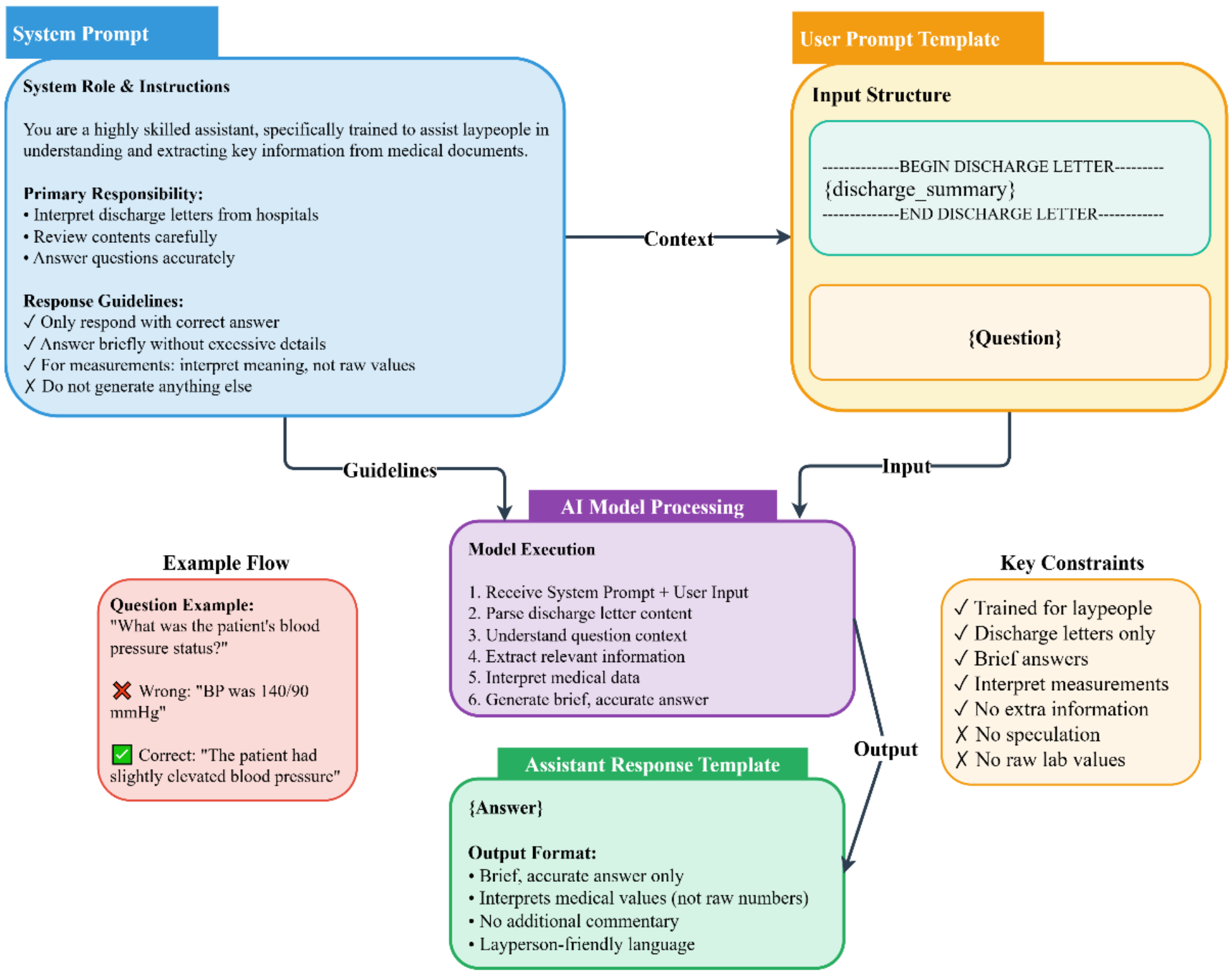

**Figure 2:** A sample de-identified clinical discharge summary and CoT prompting technique, forming the base document for QA generation.

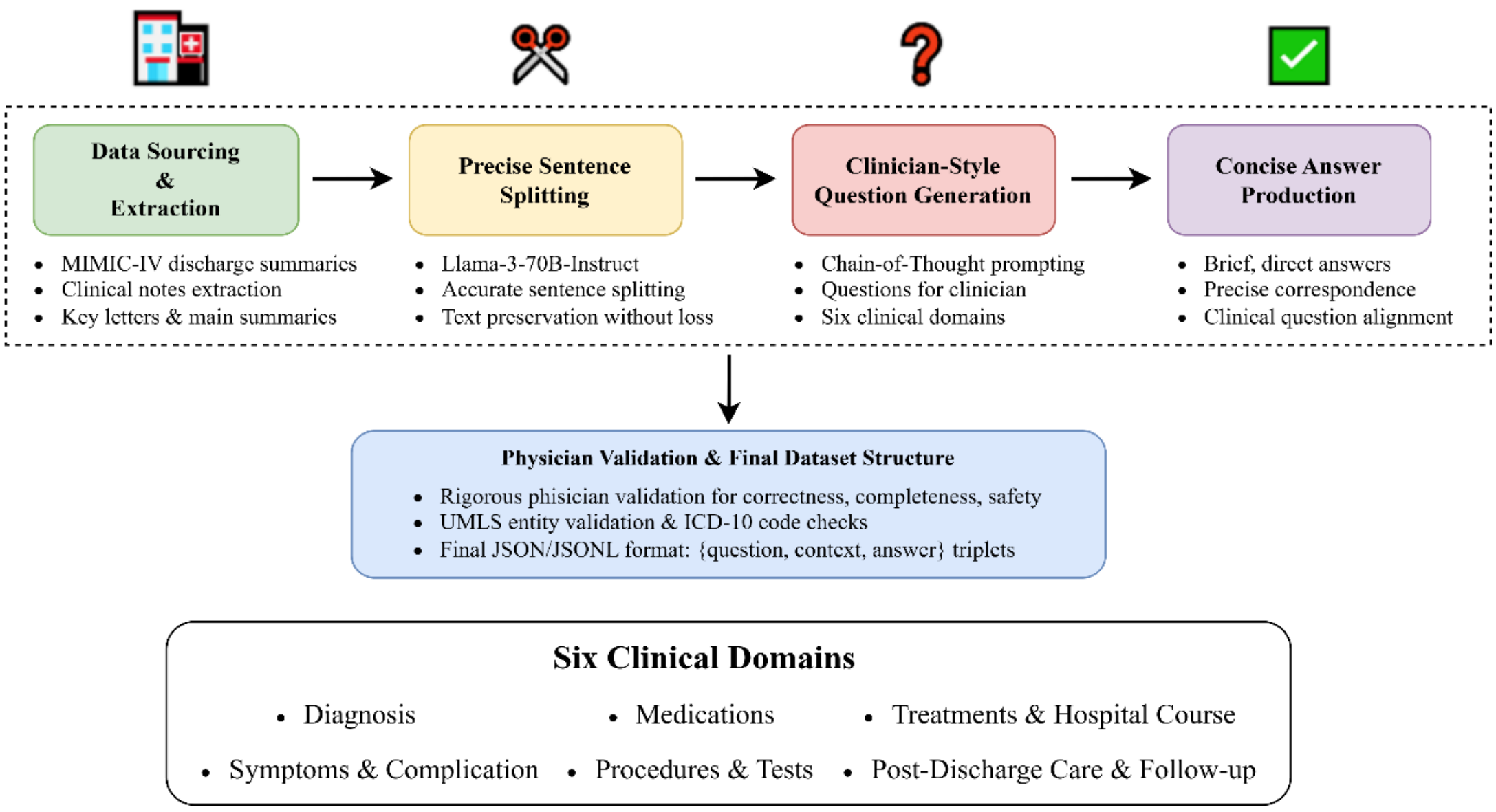

**Figure 3:** A step-by-step Chain-of-Thought data generation process from raw text to QA pairs, including sentence splitting, question generation, answer extraction, and clinician verification.

### 3.4. Dataset Overview

In total, 500 QA pairs were developed using the CoT-assisted pipeline, each grounded in real-world clinical documents and refined through expert verification. PubMedQA was not used for dataset creation or training; it was reserved solely as an external benchmark to evaluate generalization on both patient-focused and literature-based clinical reasoning tasks. Figure 4 shows the examples of the resulting QA pairs generated from the MIMIC-IV dataset.

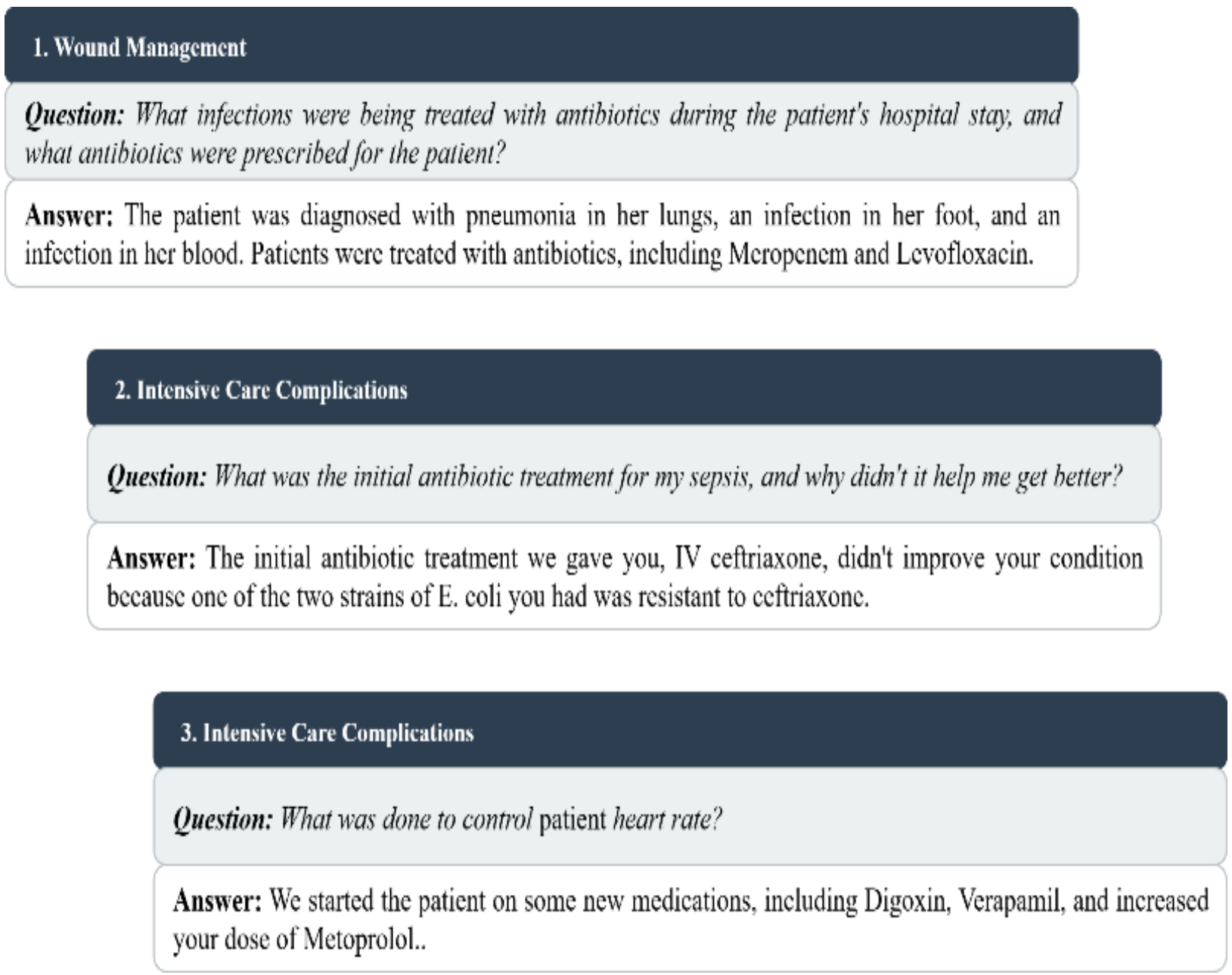

**Figure 4:** Example of QA pairs in the DocTalk-QA dataset.

### 3.5. Pretrained LLMs and Optimization

We employed six different open-source models to verify the efficacy of the strategic prompting and CoT reasoning on the DocTalk-QA and PubMedQA datasets. These enable us to examine domain-specific reasoning capabilities under consistent tuning conditions. Llama-3.1-8B-Instruct, Llama-3-8B-Instruct, and Mistral are general-purpose LLMs [64,65], whereas Llama3-Med42-8B, Meditron, and BioMistral are biomedical-specialized BioLLMs [23,66,67]. These models were subsequently fine-tuned using LoRA [68,69], a parameter-efficient fine-tuning method for efficient instruction-tuning in the medical domain. This method allowed us to quickly test various architectures by efficiently specializing large models for medical QA without requiring the entire parameter space. Each model is a causal Transformer with multi-head self-attention and gated MLPs.

We applied LoRA to all six models with the following settings: Rank (r) = 16, Alpha = 32, Dropout = 0.05, and Target Modules = q_proj, k_proj, v_proj, o_proj, gate_proj, up_proj, and down_proj. This ensures that clinical knowledge can be effectively incorporated into each model during instruction-tuning without requiring a complete fine-tuning of all parameters. To optimize the model, the learning rate was set to 2e-4 with a cosine decay schedule and a 3% warm-up to stabilize early-stage updates. A gradient accumulation of 8 was employed to simulate a higher batch size while preserving hardware efficiency, and the number of epochs was set to 10. To maximize the memory use on GPU resources, gradient checkpointing was combined with mixed precision (fp16 and bf16). The computation formula for LoRA can be expressed as:

$$y = W_0x + (\alpha/r)BAx \quad (3)$$

where x represents input features, $W_0$ denotes frozen pre-trained weights, $A \in \mathbb{R}^{dxr}$ and $B \in \mathbb{R}^{rxd}$ are trainable low-rank matrices, $\alpha$ is the scaling factor, and $r$ represents the adaptation rank.

Table 1 summarizes the original model sizes, approximate trainable LoRA parameters, and percentage of total parameters updated during instruction tuning. Across all evaluated models, only approximately 0.36% of the full model parameters were updated, demonstrating that the proposed fine-tuning strategy substantially reduces computational cost while enabling domain adaptation for clinical question answering.

**Table 1.** Parameter efficiency of LoRA-based instruction tuning across the evaluated LLMs.

| Model | Original Parameters | Trainable Parameters (LoRA) | Model Parameters Tuned (%) |
|---|---|---|---|
| Mistral-7B-Instruct-v0.1 | ~7B | ~25M | 0.36 |
| BioMistral-7B | ~7B | ~25M | 0.36 |
| Llama-3-8B-Instruct | ~8B | ~29M | 0.36 |
| Llama-3.1-8B-Instruct | ~8B | ~29M | 0.36 |
| Llama-3-Med42-8B | ~8B | ~29M | 0.36 |
| Meditron-3-8B | ~8B | ~29M | 0.36 |

### 3.6. Few-Shot and Chain-of-Thought Strategic Prompting

Strategic prompting is essential to enhance the reasoning quality of LLMs and avoid hallucination by grounding decisions in evidence from the text. In this study, we employed strategic prompting approaches that combine CoT and Few-Shot techniques. These approaches present the model with examples having reasoning chains along with input and outputs, which guide the LLM to solve the given problem in a step-by-step manner. Our workflow is structured into two stages: constructing a strategy-based clinical dataset using CoT and performing model training using Few-Shot prompting.

**Stage 1:** In the first stage, we used CoT prompting strategy to generate reasoning-aware QA pairs from pre-processed discharge summaries, shown in Figure 5 (a). This approach guided the model to simulate clinical reasoning processes, linking patients history, investigations, treatments, and outcomes, to produce interpretable and logically structured answers. The CoT design improved transparency, interpretability, and diagnostic consistency during dataset construction.

**Stage 2:** In the second stage, a few-shot (k=2) was employed for model training to improve generalization across unseen clinical questions. Two concise exemplars were provided before each test item, each comprising a discharge summary excerpt (about 4,000-6,000 tokens), a clinician-style question, and a short, chart-grounded answer, shown in Figure 5 (b). Examples were sampled only from the training split, ensuring no overlap with development or test sets. To maintain the diversity, exemplars were chosen to reflect key clinical question types, which include “final diagnosis”, “dated procedures”, “medication changes with dose, route, frequency, and rationale”, “complications”, “microbiology findings with corresponding therapies”, and “discharge disposition or follow-up care”.

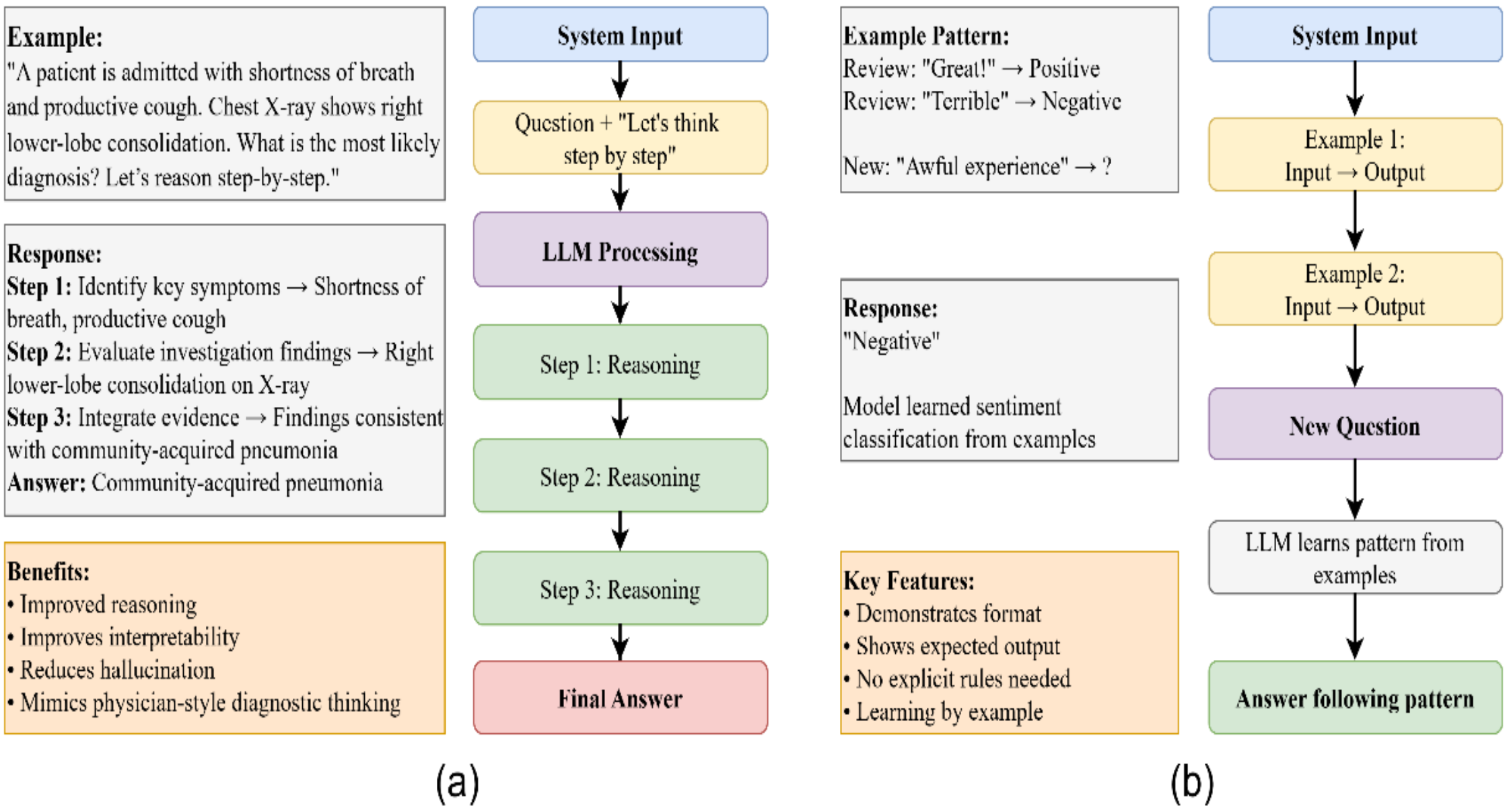

**Figure 5:** Illustration of CoT and Few-shot strategic prompting. (a) Chain-of-Thought technique for dataset construction. (b) Few-shot technique for model training.

### 3.7. Evaluation Settings

To evaluate the models during training and testing, we used LoRA-Supervised Fine-Tuning (LSFT) masked-loss strategy, with focus only on answer tokens. The loss function is defined as:

$$L_{SFT} = -\sum_i log\ p(y_i | y_{<i}, x_{clinical}) \quad (4)$$

where $y$ represents the sequence of answer tokens, $y_{<i}$ denotes the preceding token, and $x_{clinical}$ indicates the clinical input context.

To evaluate the accuracy scores of the model, we used ROUGE-1, ROUGE-2, ROUGE-L, and BERTScore-F1 performance metrics to measure the textual overlap (how closely the generated answers matched the reference answer) [70,71]. ROUGE-1 measures unigram (word-level) overlap between prediction and reference, ROUGE-2 measures biggram (two-word sequence) overlap, capturing short-range fluency, ROUGE-L measures the longest common subsequence (LCS) between prediction and reference, reflecting sentence-level structure, and BERTScore-F1 computes semantic similarity using contextual embeddings from a pre-trained BERT model, scoring how meaningfully the generated answer aligns with the reference. While UMLS-F1 is used to measure the clinical accuracy beyond the surface text similarity. These domain-specific metrics ensure that the model's outputs are not only textually fluent but also clinically valid and semantically aligned with real-world medical knowledge. UMLS-F1 evaluates concept-level correctness by comparing predicted and reference clinical entities mapped to Concept Unique Identifiers (CUIs) in the Unified Medical Language System.

## 4. Experimental Results

### 4.1. Pretrained Model Evaluation on DocTalk-QA

In this section, we have evaluated six widely used LLMs on the DocTalk-QA dataset to create a baseline before any task-specific tuning. As shown in Table 2, the comparative analysis of six different models before instruction-tuning and their performances across lexical overlap (ROUGE-1/2/L), semantic similarity (BERTScore-F1), and clinical correctness (UMLS-F1). From the comparative analysis, we can observe that Llama-3.1-8B-Instruct achieved the strongest overall performance, with the highest ROUGE-L, ROUGE-1, BERTScore-F1, and UMLS-F1 of 0.24, 0.29, 0.69, and 0.23, respectively. Close behind were the bio-LLMs (Meditron-3-8B and Llama3-Med42-8B), which obtained comparable semantic similarity scores (BERTScore-F1=0.66-0.68) and moderate clinical correctness scores (UMLS-F1=0.20-0.22). Moreover, the general-LLMs (Llama-3-8B-Instruct and Mistral-7B-Inst-v0.1) performed reasonably but showed limited clinical grounding. On the other hand, BioMistral-7B, despite being domain-specialized, performed poorly across all metrics, indicating difficulty with answer-only clinical QA without task supervision.

**Table 2.** Baseline performance of pretrained LLMs on the DocTalk-QA

| Model | ROUGE-L | ROUGE-1 | ROUGE-2 | BERTScore-F1 | UMLS-F1 |
|---|---|---|---|---|---|
| **Llama-3.1-8B-Instruct** | **0.24** | **0.29** | **0.10** | **0.69** | **0.23** |
| Meditron-3-8B | 0.21 | 0.27 | 0.08 | 0.66 | 0.22 |
| Llama-3-Med42-8B | 0.21 | 0.26 | 0.08 | 0.67 | 0.21 |
| Llama-3-8B-Instruct | 0.20 | 0.26 | 0.08 | 0.67 | 0.21 |
| Mistral-7B-Instruct-v0.1 | 0.21 | 0.26 | 0.08 | 0.67 | 0.20 |
| BioMistral-7B | 0.12 | 0.16 | 0.03 | 0.58 | 0.11 |

We can observe two clear patterns from Table 2: first, among pretrained models, semantic similarity scores remain moderate (BERTScore-F1=0.66-0.69), which shows that answers are generally well-formed but not deeply grounded in the clinical context, and second, all baseline models did not perform well in terms of clinical fidelity (UMLS-F1=0.11-0.23). Therefore, these baselines require task-specific supervision, even though they can generate concise and coherent outputs.

### 4.2. Instruction-Tuned Model Performance on DocTalk-QA

In this section, we first fine-tuned all six LLMs using the LoRA-based instruction-tuning configuration and then evaluated the tuned models on the DocTalk-QA dataset, shown in Table 3. After fine-tuning, the models produced substantial improvements across all metrics, especially in clinical correctness scores, demonstrating the effectiveness of DocTalk's structured supervision. In terms of lexical overlap, semantic similarity, and clinical grounding, the performance consistently increased across all models. The promising overall performance was achieved by bio-LLM (Llama-3-Med42-8B), which improved ROUGE-L from 0.21 to 0.40, BERTScore-F1 from 0.67 to 0.76, and UMLS-F1 from 0.21 to 0.49. Moreover, comparable gains for general-LLMs were observed, +0.16 ROUGE-L, +0.06 BERTScore, and +0.24 UMLS-F1 for Llama-3.1-8B-Instruct and +0.17 ROUGE-L, +0.07 BERTScore, and +0.26 UMLS-F1 for Llama-3-8B-Instruct. At the same time, other bio-LLMs, such as Meditron-3-8B and BioMistral-7B, also improved considerably in ROUGE-L (+0.17 to 0.23) and UMLS-F1 (+0.25 to 0.30). The average scores of the finetuned models improved by 0.18 for ROUGE-L, 0.09 for BERTScore, and 0.26 for UMLS-F1, demonstrating that the domain adaptation enhances clinical correctness more than lexical similarity.

**Table 3.** Performance of Instruction-tuned LLMs on DocTalk-QA

| Model | ROUGE-L | ROUGE-1 | ROUGE-2 | BERTScore-F1 | UMLS-F1 |
|---|---|---|---|---|---|
| **Llama-3-Med42-8B** | **0.40** | **0.44** | **0.23** | **0.76** | **0.49** |
| Llama-3.1-8B-Instruct | 0.40 | 0.44 | 0.21 | 0.75 | 0.47 |
| Llama-3-8B-Instruct | 0.37 | 0.41 | 0.22 | 0.74 | 0.47 |
| Meditron-3-8B | 0.38 | 0.43 | 0.21 | 0.75 | 0.47 |
| Mistral-7B-Instruct-v0.1 | 0.37 | 0.42 | 0.20 | 0.75 | 0.44 |
| BioMistral-7B | 0.35 | 0.39 | 0.18 | 0.74 | 0.41 |

### 4.3. Performance Evaluation on PubMedQA

To assess whether the DocTalk instruction-tuning strategy generalizes beyond discharge-summary question answering, we further evaluated two representative models on the external PubMedQA benchmark. The models were selected based on their performance on DocTalk-QA: Llama-3.1-8B-Instruct, which achieved the strongest pretrained baseline performance, and Llama-3-Med42-8B, which achieved the best overall results after instruction tuning. To examine the effect of DocTalk-based adaptation, both models were evaluated on PubMedQA before and after instruction tuning. PubMedQA requires reasoning over biomedical research abstracts and includes Yes/No/Maybe answers, making it a significantly different challenge from DocTalk's answer-only clinical format. Although the full benchmark includes Yes/No/Maybe responses, the confusion-matrix analysis presented here focuses on the binary Yes/No classification setting. The evaluation set is markedly imbalanced: [N_yes] "Yes" and [N_no] "No" instances (92.8% positive). A majority-class predictor that always outputs "Yes" therefore attains 0.928 accuracy, which limits the informativeness of raw accuracy on this benchmark. We accordingly adopt the Matthews correlation coefficient (MCC) as the primary discrimination metric, since it accounts for performance across both classes, and report accuracy, precision, recall, and F1 as secondary measures. Figure 6 illustrates the confusion matrices for the four evaluation conditions: Llama-3.1-8B-Instruct baseline, Llama-3-Med42-8B baseline, Llama-3.1-8B-Instruct instruction-tuned, and Llama-3-Med42-8B instruction-tuned. Table 4 summarizes the corresponding classification results.

The baseline models yielded comparable results on PubMedQA (accuracy = 0.931 for Llama-3.1-8B-Instruct 0.935 for Llama-3-Med42-8B). Both models were improved after DocTalk-based instruction tuning, but with dramatically different magnitude of improvement. Llama-3.1-8B-Instruct showed the largest gain, improving the accuracy from 0.931 to 0.977, recall from 0.930 to 0.983, F1-score from 0.962 to 0.987, and MCC from 0.661 to 0.839. This suggests that discrimination and class-balanced predictive reliability have achieved a significant improvement. The instruction-tuned Llama-3-Med42-8B also had an increase in performance compared to baseline, with accuracy increasing from 0.935 to 0.943, F1-score from 0.964 to 0.968, and MCC from 0.675 to 0.707. Although the gain was more modest than that of Llama-3.1-8B-Instruct, the result confirms that DocTalk-based instruction tuning does not only improve performance on the in-domain DocTalk-QA task, but can also transfer positively to biomedical abstract-level reasoning.

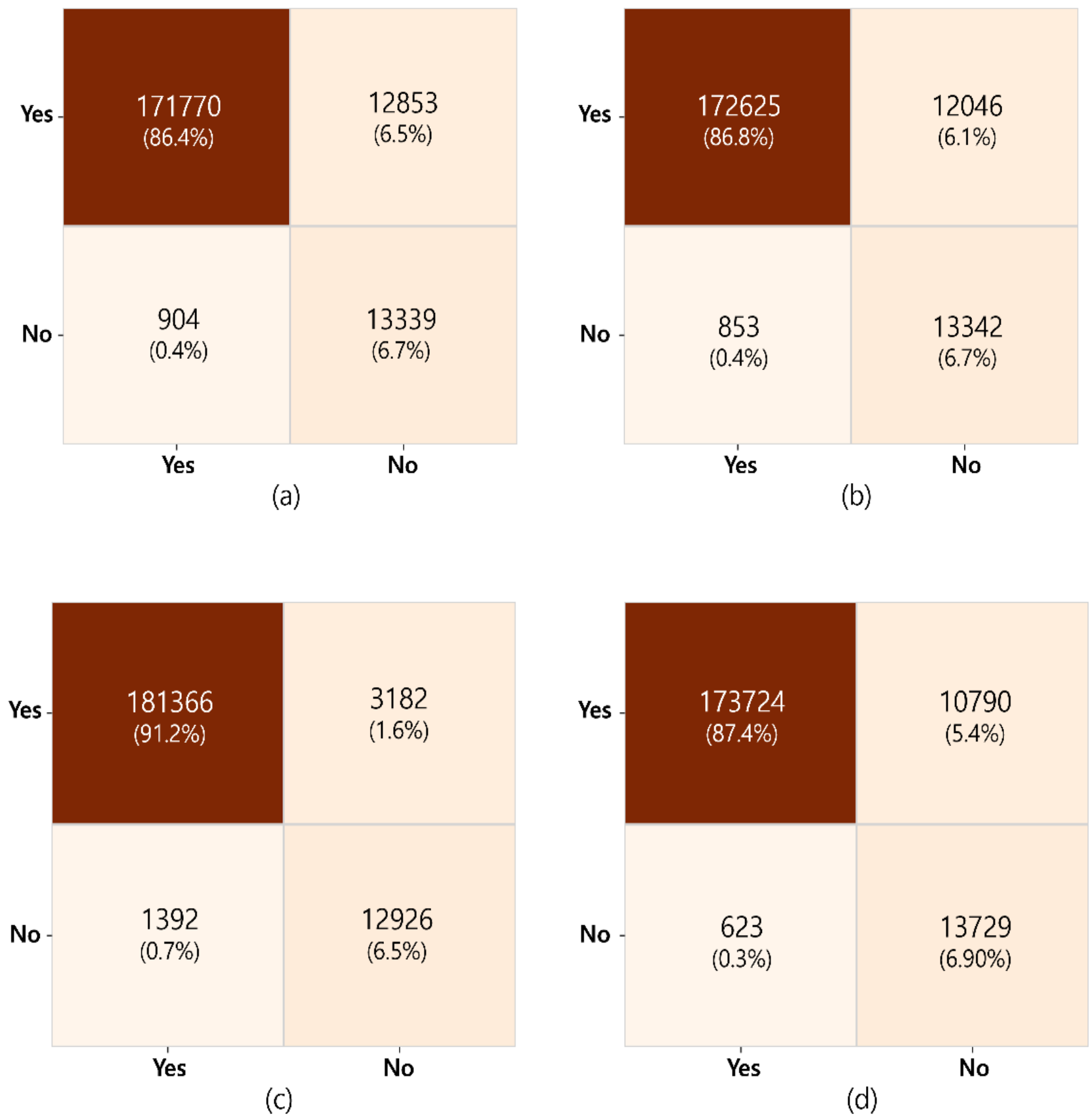

**Figure 6:** The confusion matrices on the PubMedQA benchmark for the selected models before and after instruction tuning. (a) Llama-3.1-8B-Instruct baseline. (b) Llama-3-Med42-8B baseline. (c) Llama-3.1-8B-Instruct instruction-tuned (d) Llama-3-Med42-8B instruction-tuned.

**Table 4.** Comparison performance of baseline and instruction-tuned LLMs on PubMedQA.

| Model | Accuracy | Precision | Recall | F1-Score | MCC |
|---|---|---|---|---|---|
| Llama-3.1-8B-Instruct (Baseline) | 0.931 | 0.995 | 0.930 | 0.962 | 0.661 |
| Llama-3-Med42-8B (Baseline) | 0.935 | 0.995 | 0.935 | 0.964 | 0.675 |
| **Llama-3.1-8B-Instruct (Instruction-Tuned)** | **0.977** | 0.992 | **0.983** | **0.988** | **0.839** |
| Llama-3-Med42-8B (Instruction-Tuned) | 0.943 | **0.996** | 0.942 | 0.968 | 0.707 |

These findings reveal two important patterns. First, the DocTalk instruction-tuning framework provides measurable generalization benefits on PubMedQA for both selected models. Second, the model that performs best on DocTalk-QA after instruction tuning, Llama-3-Med42-8B, is not necessarily the strongest model on the external PubMedQA benchmark. This suggests that in-domain clinical QA performance and out-of-domain biomedical reasoning performance are related but not identical, and both should be considered when assessing the practical utility of medical LLMs. To further contextualize the external benchmark performance, the best-performing DocTalk configuration was compared with previously reported PubMedQA systems in Table 5.

**Table 5.** Comparison performance of DocTalk with other LLMs on PubMedQA.

| Model | Accuracy |
|---|---|
| **DocTalk** | **0.977** |
| Gyan-4.3 | 0.871 |
| SM70 | 0.773 |
| Human Expert | 0.780 |
| BioGPT-Large | 0.630 |
| FLAN-PaLM | 0.790 |
| Med-PaLM 2 | 0.750 |
| GPT-4 (5-shot) | 0.752 |
| GPT-4-base (5-shot) | 0.594 |
| Alpaca-7b | 0.380 |
| LLaMA-7b | 0.594 |

## 5. Discussion

In this study, to demonstrate the real-world applicability of our instruction-tuned biomedical QA system, we developed a fully functional web-based platform, DocTalk, enabling clinicians and researchers to interact with large language models directly using electronic health records (EHRs). The system supports secure login, patient report selection, and natural-language questioning, simulating real clinical workflows (Supplementary Information, Figure S1). Unlike most prior research that relies on simulated datasets or benchmark QA pairs separated from practical clinical use [22, 72-74], our work takes a grounded, data-centric approach by curating questions directly from authentic clinical narratives. Our methodology ensures high reliability of the original chart content and reflects realistic physician information needs. We contributed in three major areas. First, we introduced a rigorous dataset construction pipeline that generates QA pairs from atomic sentence-level decomposition with concise CoT prompting, followed by manual quality assurance. This careful design directly addresses weaknesses in existing datasets, such as BioASQ, that often lack clinical relevance, structure, or factual precision [73]. Second, we proposed and implemented a unified prompting strategy incorporating system-grounding, few-shot in-context examples, and answer-only supervision, fine-tuned via LoRA. This resulted in significant improvements across both textual and clinical metrics. Specifically, our instruction-tuned Llama-3-Med42-8B achieved performance that outstripped a strong baseline, especially on clinically meaningful measures like UMLS-F1. Compared with evaluation strategies that rely mainly on surface-level text similarity, our inclusion of clinically grounded metrics provides a more medically meaningful assessment of QA quality. Unlike models such as BioGPT [74] or Med-PaLM [35], which prioritize large-scale pretraining or multi-choice reasoning, DocTalk is optimized for generating short, precise, and chart-grounded answers, crucial for real-time clinical use. Third, we demonstrated that our approach is feasible in real-time settings with limited resources. The use of parameter-efficient tuning (LoRA) is more scalable, and our instruction tuning method is concise enough to be clinically relevant and suitable for point-of-care integration. Previous approaches typically focus on benchmark scores or exam-style QA in a practical manner [35], while our pipeline focuses on interpretability, brevity, and chart fidelity, which are all important for real-world clinical decision support systems.

The external PubMedQA evaluation provides further insight into the generalization capacity of the DocTalk instruction-tuning strategy. Both selected models achieved higher performance after instruction tuning, demonstrating that supervision derived from clinician-oriented discharge-summary QA can transfer beyond the original DocTalk-QA task. The improvement was particularly substantial for Llama-3.1-8B-Instruct, whose accuracy increased from 0.931 to 0.977, accompanied by clear gains in recall, F1-score, and MCC. Llama-3-Med42-8B also improved after tuning, although by a smaller margin. These findings suggest that DocTalk-based adaptation strengthens biomedical reasoning in a model-dependent manner rather than producing uniform gains across architectures. At the same time, the results show that the best-performing model on the in-domain DocTalk-QA dataset is not necessarily the best-performing model on the external PubMedQA benchmark. This distinction is important for clinical AI development, as strong performance on curated clinical QA should be complemented by cross-benchmark validation to assess whether the learned knowledge transfers effectively to broader biomedical reasoning tasks.

However, our work has some limitations. First, the QA pairs used for training and evaluation originally stemmed mainly from discharge summaries coming from a single healthcare organization. This raises concerns about the generalizability of our model across different hospital systems, clinical subspecialties, and documentation styles. Furthermore, although our instruction-tuned model shows lower hallucinations and better factual fidelity, it makes errors in some complex scenarios combining multiple comorbidities or sophisticated clinical reasoning. One disadvantage is the model's exclusive emphasis on single-turn question answering. Second, real-world clinical conversations are multi-turn, which can lead to challenges of contextual recall, follow-up queries, and reasoning from temporally scattered information. Adding multi-hop and dialog-based reasoning capabilities presents a critical avenue for future enhancements. Third, there is the fact that the existing framework is based exclusively on English for development and evaluation. Due to the linguistic variation within the world's healthcare systems, future studies should examine cross-lingual adaptation and performance in non-English clinical data corpora. While out-of-distribution generalization was partially addressed, the nature of the dynamic biomedical knowledge requires continual refinement and validation in order to ensure further long-term robustness and relevance.

Future iterations will focus on extending the DocTalk platform by providing advanced use cases, for example, summarizing patient data, generating diagnostic recommendations, or tracking temporal reports. Moving on to safe and effective deployment with real-world implementations, we will also need feedback from medical professionals and undertake prospective clinical trials at scale.

## 6. Conclusion

We present DocTalk, a clinically grounded, instruction-tuned QA system with the aim of improving chart comprehension and helping physicians in real-world EHR review workflows. Our methodology balances efficiency with medical relevance by generating high-fidelity QA pairs from de-identified discharge summaries and fine-tuning open-source LLMs via a lightweight LoRA-based framework. Specifically, our models, such as Llama-3-Med42-8B, outperform others not only on text-based metrics but also on clinically aligned metrics.

Assessment on the external PubMedQA benchmark further showed that DocTalk-based instruction tuning can improve biomedical reasoning beyond the original clinical QA setting, although the magnitude of transfer varies across model architectures. More broadly, we have also deployed our QA system as part of a web platform, a platform that also allows interactive clinician engagement, resulting in a system that has proven real-world useful and accessible. Our contributions, ranging from data generation, instruction tuning, innovation for evaluation, and deployment, provide a scalable solution for embedding LLMs in healthcare applications.